\documentclass[conference]{IEEEtran}

\usepackage{cite}
\usepackage{float}
\usepackage{dirtytalk}
\usepackage[pdftex]{graphicx}
\graphicspath{images/}
\usepackage{amsmath}
\usepackage{subcaption}
\usepackage{multirow}

\DeclareMathOperator*{\argmax}{argmax}

\usepackage{url}

\begin{document}
\title{Image Captioning and Classification of Dangerous Situations}

\author{\IEEEauthorblockN{Octavio Arriaga}
        \IEEEauthorblockA{Hochschule Bonn-Rhein-Sieg \\
        Sankt Augustin Germany  \\
        Email: octavio.arriaga@smail.inf.h-brs.de}
        \and
        \IEEEauthorblockN{Paul G. Pl\"oger}
        \IEEEauthorblockA{Hochschule Bonn-Rhein-Sieg \\ 
        Sankt Augustin Germany \\
        Email: paul.ploeger@h-brs.de}
        \and
        \IEEEauthorblockN{Matias Valdenegro}
        \IEEEauthorblockA{ Heriot-Watt University \\
        Edinburgh, UK \\
        Email: m.valdenegro@hw.ac.uk}}

    \maketitle

        \begin{abstract}
        Current robot platforms are being employed to collaborate with humans in a wide range of domestic and industrial tasks.
        These environments require autonomous systems that are able to classify and communicate anomalous situations such as fires, injured persons, car accidents; or generally, any potentially dangerous situation for humans.
        In this paper we introduce an anomaly detection dataset for the purpose of robot applications as well as the design and implementation of a deep learning architecture that classifies and describes dangerous situations using only a single image as input.
        We report a classification accuracy of 97 \% and METEOR score of 16.2. 
        We will make the dataset publicly available after this paper is accepted.
        \end{abstract}

        \IEEEpeerreviewmaketitle

        \section{Introduction}
        Autonomous systems are currently being employed to conduct households activities, as personnel in healthcare institutions and as general patrolling systems.
        All of these circumstances require an agent that is able to classify and communicate irregularities and anomalous situations.
        In this paper we consider that an image is an anomaly if it contains one of the following classes: broken windows, injured people, fights, explosions, car accidents, fire accidents, guns and domestic violence.
        Unfortunately, the current robot platforms are unable to attend these demands since building a robust anomaly detection system would require an integration of multiple components for all specific cases in which a situation is potentially dangerous.
        Consequently, most of the current anomaly detection algorithms are built for specific test cases such as fight detection \cite{Demarty2015}.
        Therefore, we opted for a machine learning (ML) approach, in which we could learn the features that consider an image an anomaly instead of trying to account for all possible situations.
        However, a problem regarding ML techniques applied to anomaly detection is the limited amount of legally available datasets that contain a set of real-world anomalies that we would like to classify and caption.
        In respond to all this issues we introduced an anomaly detection dataset consisting of more than 1000 captioned images, as well as the design and implementation of a deep learning architecture that is not only able to classify if an image is an anomaly or not but also provides a full description of such anomaly.
        We argue that a complete sentence brings more information about the situation than simply providing a classification of the anomaly such as broken windows, injured people or car accidents.
        For example, an image that is classified as an anomaly but only reports the class \say{injury} brings little information about the subject to attend; however, a system that classifies the situation as an anomaly but also communicates \say{there is a woman with blood on the floor} could prove more beneficial in order to address such situation.
        The anomalies presented in our dataset were specifically selected in order to further develop robot activities in the areas of: domestic-services and patrolling systems.
        Using our architecture the robot would now be able to communicate in natural language what is perceives.
        A robot with this ability could be deployed in hospitals, houses, supermarkets, concerts and sport venues as either a mobile or a static platform.
        Thus, the main contributions can be summarized as: 1) we created a new anomaly detection dataset for the purpose of robotic platforms; 2) we designed and implemented a deep learning architecture that classifies and describes a dangerous situation using a single image as input.

        \section{Related Work}

        \subsection{Image Captioning}
        Making a machine classify and describe the environment it perceives is one of the hardest open problems in artificial intelligence (AI). This problem combines two of the most fundamental areas in AI: computer vision and natural language processing \cite{Lebret2015}.
        However, deep learning methods have been able to overcome long lasting problems in AI and machine learning, and they have been successfully applied to the problem of understanding a scene using image captioning \cite{Vinyals2016b}.
        The winner implementation of the COCO-2015 Image Captioning challenge was originally proposed in \cite{Vinyals2015} and later modified by the same authors on \cite{Vinyals2016b}.
        One of its main advantages over previous image captioning models \cite{Fang2015} \cite{Lebret2015} is that this architecture is trainable end-to-end; consequently, it alleviates slow performances and complicated pipelines. 
        The overall approach of this model is that given an image $I$, they train a probabilistic language model $p$ with parameters $\theta$, so that it maximizes the likelihood $p(S|I;\theta)$, where $S$ is a sequence of words $S = \{S_1,S_2,...\}$.
        This is expressed as:
            \begin{equation} \label{eq:max_likelihood}
                \theta^{*} = \argmax_{\theta} \sum_{I,S} \log p(S|I;\theta)
            \end{equation}
        We know that given a log joint probability distribution $\log p(S|I;\theta)$ we can apply the chain rule of joint probabilities over the words $S = \{S_0,...,S_N\}$.
        Assuming that this sentence contains $N$ words this results on following equation:
            \begin{equation} \label{eq:log_probability}
                \log p(S|I;\theta) = \sum_{t=0}^{N} \log p(S_t|I,S_0,...,S_{t-1};\theta)
            \end{equation}
        In the equation above the product between the probabilities turns into a sum since we are taking the logarithm of the likelihood.
        Therefore, our optimization task gets reduced to maximize the sum of these log probabilities given in the equation above \cite{Vinyals2015}.

        The novelty of this approach is that the language model is given by recurrent neural network (RNN) and the image encoding is performed by a CNN; consequently, it uses the weights of the networks as the parameters for which we would like to maximize the likelihood \cite{Sutskever2014sequence}.
        An RNN fits naturally to this approach since we can encode information from a variable number of previous given words into a hidden state in the network \cite{Vinyals2015}.
        The model uses as RNN a long short-term memory network (LSTM) to decode it into a sentence $S$.
        This joint model receives the name \textit{neural image caption} (NIC).

        The image encoding was done using GoogLeNet without its last classification layer. This penultimate layer outputs a vector of $2048$ dimensions which corresponds to the relevant features of the image.
        Each word is represented using a vector of dimensions equal to the vocabulary size using a one-hot-encoding.
        Once they have a numerical representation of words and images, the authors proceeded by embedding words and images into a vector space of the same dimensions.
        They do this by adding a fully connected layer that takes as input the encoded image and maps it into a smaller dimension D.
        Similarly, they construct another fully connected layer that transforms every one hot encoding into a vector of the same dimension D.
        Finally, a LSTM networks is initialized with the embedded image features and a special word-token that indicates the beginning of the sentence. 
        This LSTM network calculates at every time step using a softmax activation function, a vector which dimension equals the number of words in the vocabulary.
    Therefore, this last vector assigns probability to each word at every time-step of the LSTM.
        \begin{figure}
                \centering
                \includegraphics[width=0.9\linewidth]{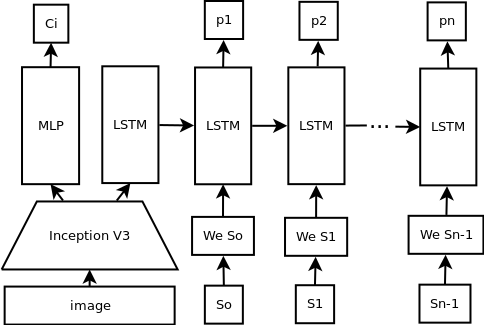}
                \caption{Our proposed model for image captioning and classification.}
                \label{fig:image_captioning_model_show_and_tell}
        \end{figure}

        The equations that describe the entire CNN-LSTM pipeline are written below.
        \begin{align}
            \begin{split}
                    x_{-1} ={}& W_I \text{CNN}(I)
            \end{split} \\
            \begin{split}
                    x_t ={}& W_e S_t, t \in \{0,...,N-1\}
            \end{split} \\
            \begin{split}
                    p_{t+1} ={}& \text{LSTM}(x_t), t \in \{0,...,N-1\}
            \end{split} \\
        \end{align}
        The matrices $W_I$ and $W_S$ represent the fully connected layers with a linear activation function. 
        In this setup the image $I$ is only taken as input once. 
        The authors in \cite{Vinyals2015} had empirically tested that giving the image as input at every time step yields worse results.
        The LSTM network was trained to predict the next word in a sentence, given the image and the previous hidden state.
        Consequently the model is better expressed by unrolling the LSTM network; thus, making a copy for every word in the sentence.

        \subsection{Anomaly detection}
        There exist a limited amount of datasets that contain anomalous situations. We refer to anomalous situations as events that could prove potentially dangerous for humans.
        The existing datasets suffer from some deficiencies regarding our application; mainly: none of them are captioned, they only provide a limited amount of anomalies and some of them are not easily accessible.
        Specifically, the authors from \cite{Nievas2011} presented a dataset that only contain fights between hockey players, and the authors from  \cite{Demarty2015} created a dataset which is made of violent scenes from movies; however, this dataset can't be easily distributed due to copyright infringement.

        \section{Anomaly Detection Dataset}
        We created a dataset that contains 1008 captioned images which only correspond to anomalies (AD dataset).
        We consider that an image is an anomaly if it contains one of the following classes: broken windows, injured people, fights, car accidents, fire accidents , guns and domestic violence.
        These classes were specifically selected in order to develop robot activities such as: patrolling or domestic-services.

       All images from our dataset were selected from flickr using all creative commons license which allow us distribute the dataset under certain conditions such as: attributing the original authors and a non-commercial agreement.
        All the images were captioned by 20 different persons, who were asked to follow the next set of instructions in order to caption an image: \\
    {\small
    \begin{itemize}
        \item Write a single English sentence for each image.
        \item The sentences have to be in present or present continuous tense. Present continuous tense: Present tense of the verb \say{to be} plus the present participle (-ing form) of a verb i.e. \say{a man is laying on the ground, while two paramedics are assisting him}.
        \item Write primarily about the accident/incident/anomaly in the image.
        \item When possible be explicit with the number of persons in the image.
        \item The sentence should not contain any digits (i.e. 1 2 3 ... 9).
        \item Use written numbers instead of digits (i.e. 'one' 'two' ... 'nine').
        \item When appropriate be explicit with the gender (man, woman).
        \item There is no limitation in the length of the sentence. However, it is advised to use between 7 to 18 words per sentence.
    \end{itemize}
    }

    \begin{figure}[H]
    \begin{center}
        \includegraphics[scale=.20]{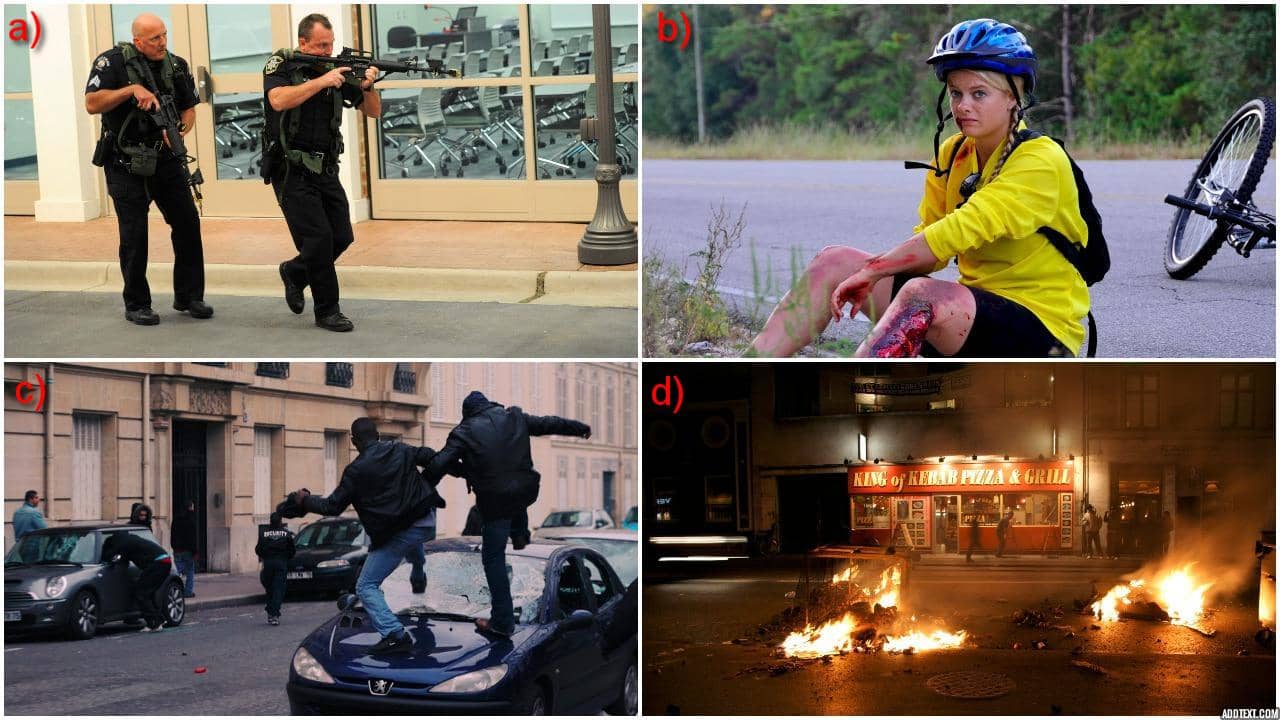}
    \end{center}
    \caption{Images from our anomaly detection dataset. The captions used for these images are: a) ``two policemen carry rifles on the street, while one of them is aiming at something`` b) ``a woman with an injured leg is sitting on the side of the road next to her bicycle`` c) ``two men kicking violently the windshield of a car`` d) ``there is something burning in front of a shop.``}
    \label{fig:add_sample}
    \end{figure}

        \section{Model}
        Our model presents shown in Fig. \ref{fig:image_captioning_model_show_and_tell} extends the image captioning in model presented in \cite{Vinyals2016b} by introducing a classification module.
        We trained our model with the combination of both the IAPR-2012 dataset \cite{Grubinger2006} and our AD dataset.
        The IAPR-2012 dataset consists of 20,000 captioned images.
        It is free of any copyright restrictions; therefore it can be easily distributed, and it contains diverse human actions.
        For example, the dataset contains images that describe human activities such as pushing, drinking, celebrating, but also contains examples of wildlife, city pictures and landscapes.

        We preprocessed our captions by removing all non-alphanumeric symbols, converting all sentences to lowercase and removing all words which had a frequency less than 3 for both datasets.
        We then created a one-hot encoding representation of every word.
        The vocabulary size for the IAPR-ADD dataset was of 1078 words.
        One main difference in our preprocessing step with regards to the other captioning models is that we discarded all captions that were longer than 14.
        This was done since we are only interested in developing sentences that are able to describe anomalous situations in a concise manner.
        After preprocessing all the target sentences we proceeded to preprocess all the images.
        This step consisted of passing all images through a CNN without it's last fully connected layer.
        For this purpose we used as CNN the Inception-V3 network, which corresponds to the third incarnation of GoogLeNet network \cite{Szegedy2014}.

        This particular trained Inception V3 network obtains top-5 error of 7.3 \% on ImageNet \cite{Chollet2015}.
        The current implementations of image captioning models use a beam search in order to refine their sentences. 
        This step could prove computationally expensive for any real-time implementation.
        The original input to the CNN ($299 \times 299$ pixels) got encoded in a feature vector of 2048 dimensions.

       Our final architecture used for the image captioning module has 512 LSTM neurons and an image embedding of 512 dimensions.
       Our classification module consisted of a multi-layer perceptron (MLP) with two hidden layers, ReLUs and dropout and it provided a binary classification between normal situations and anomalies.
        We split our dataset using 80 \% of all images for training and validation, and the 20 \% left as a test set.

        \section{Metrics}
        The standard metrics used in image captioning are BLEU \cite{Papineni2002} and METEOR \cite{Banerjee2005}.
        However, there has been criticism regarding BLEU since it has shown that it correlates poorly with human judgment in comparison to METEOR \cite{Banerjee2005} \cite{Vedantam2015} \cite{KelvinXu2015} \cite{Johnson2015}.
        METEOR calculates its score by using 3 matching components which are processed in the following order: exact, stem and synonym \cite{Banerjee2005}.
        These components make an injective mapping between both strings.
        The exact component maps unigrams that are identical, the stem component maps unigram that have the same word-stem; also called root form, and the synonym component maps two unigrams if they are synonyms.
        Since mappings are not unique METEOR always tries to maximize the number of matches.
        In the case in which two mappings have the same number of matches, METEOR selects the one that has the least amount of crosses.
        Crosses can be intuitively understood by aligning reference sentence and the machine generated sentence one above the other one, and drawing straight lines between the matched words.
        The number of crosses will then be the number of intersections between the matching connections \cite{Banerjee2005}. 
        After running the matching modules we calculate the unigram precision (P): ratio between the number of unigrams matched in the machine generated sentence and the total number of unigrams in the machine generated sentence.
        Also, the unigram recall (R) is calculated: ratio between the number of unigrams matched in the machine generated sentence and the total number of unigrams in the reference sentence.
        Then, we calculated a weighted F-score that puts more weight on the recall. In \cite{Banerjee2005} they propose using the harmonic mean of $9R$ and $P$.

            \begin{equation}
                F_{\text{score}} = \frac{10PR}{R+9P} 
            \end{equation}

        METEOR also calculates a penalty value:

            \begin{equation}
                \text{Penalty} = 0.5* \Big( \frac{\text{chunks}}{\text{unigrams matched}} \Big)^3
            \end{equation}

        Where chunks refers to the number of adjacent positions in the machine generated sentence that are mapped also to adjacent positions in the reference sentence.
        In the extreme case where the computer generated sentence is the same as the reference frame we will have only one chunk. 
        The penalty value increases when the number of chunks increases up to a maximum of $0.5$. The penalty lower bound is dictated by the number of unigram matches \cite{Banerjee2005}.
        The METEOR score is then calculated as follows:
            \begin{equation}
                \text{Score} = \text{F-score}*(1-\text{Penalty})
            \end{equation}

        \section{Results}
\begin{figure}[H]
    \captionsetup[subfigure]{}
		\centering
		\begin{subfigure}[b]{0.32\textwidth}
			\includegraphics[scale=.057]{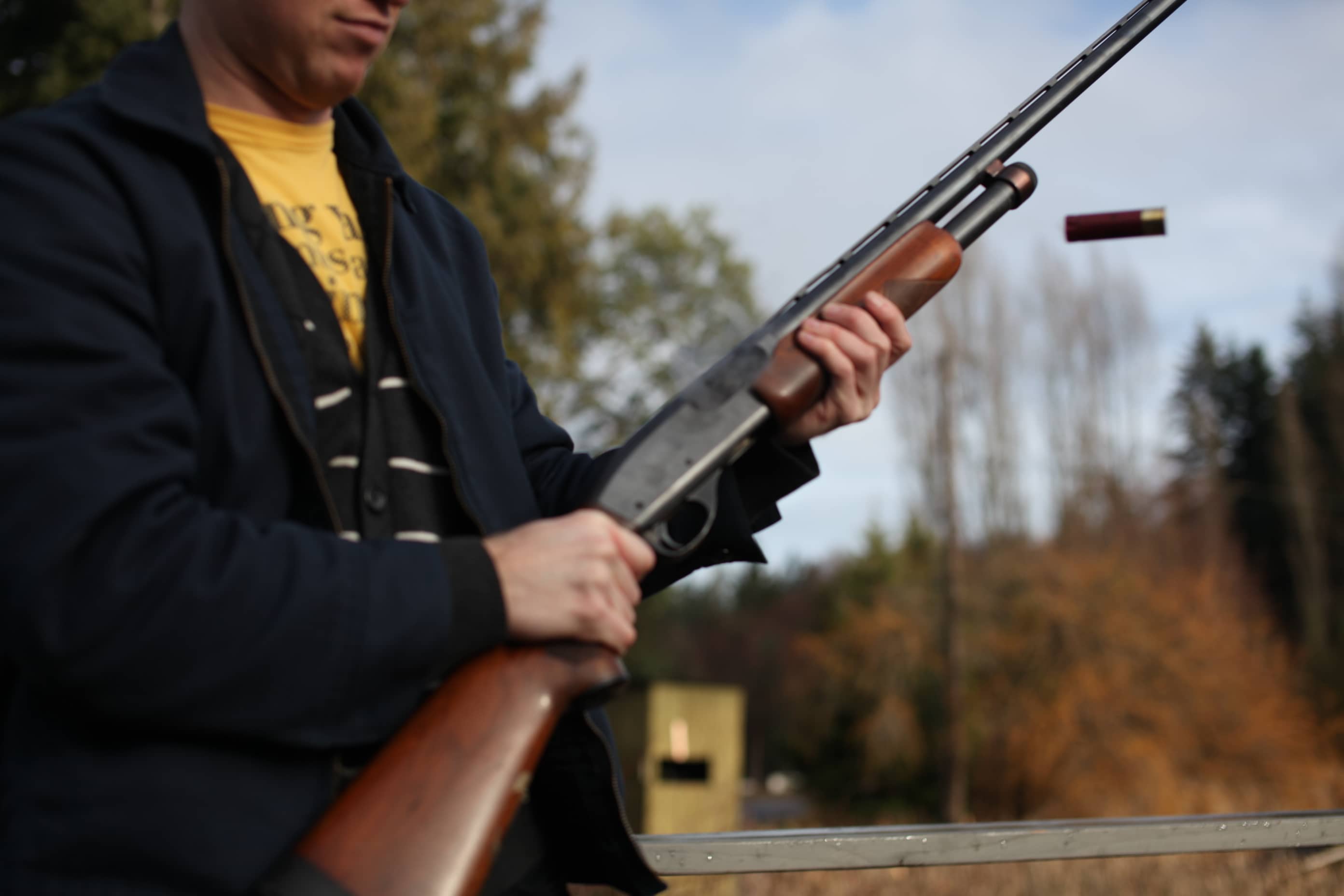}
            \caption{\say{a man is holding a gun}}
		\end{subfigure}
		\begin{subfigure}[b]{0.32\textwidth}
			\includegraphics[scale=.18]{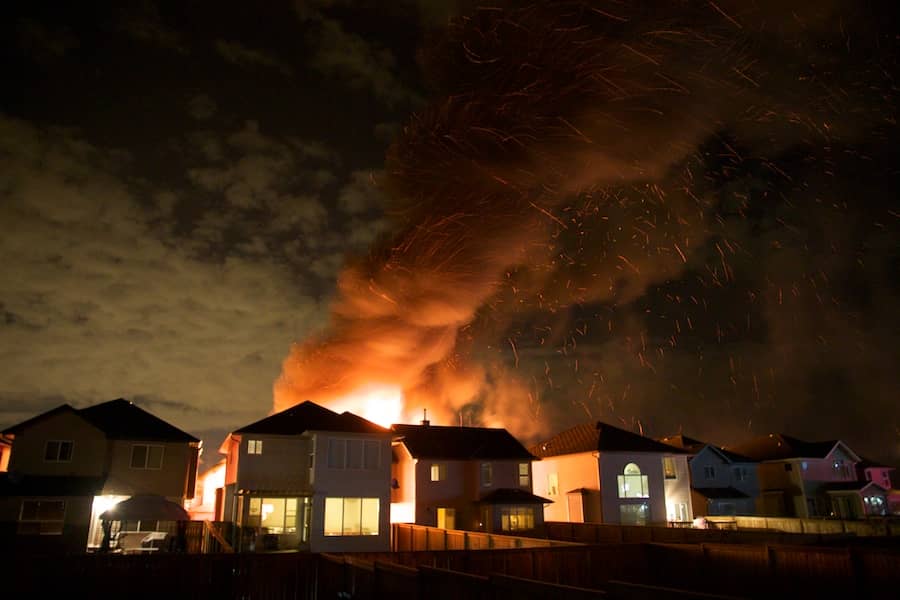}
            \caption{\say{a house is burning}}
		\end{subfigure}
		\begin{subfigure}[b]{0.32\textwidth}
			\includegraphics[scale=.11]{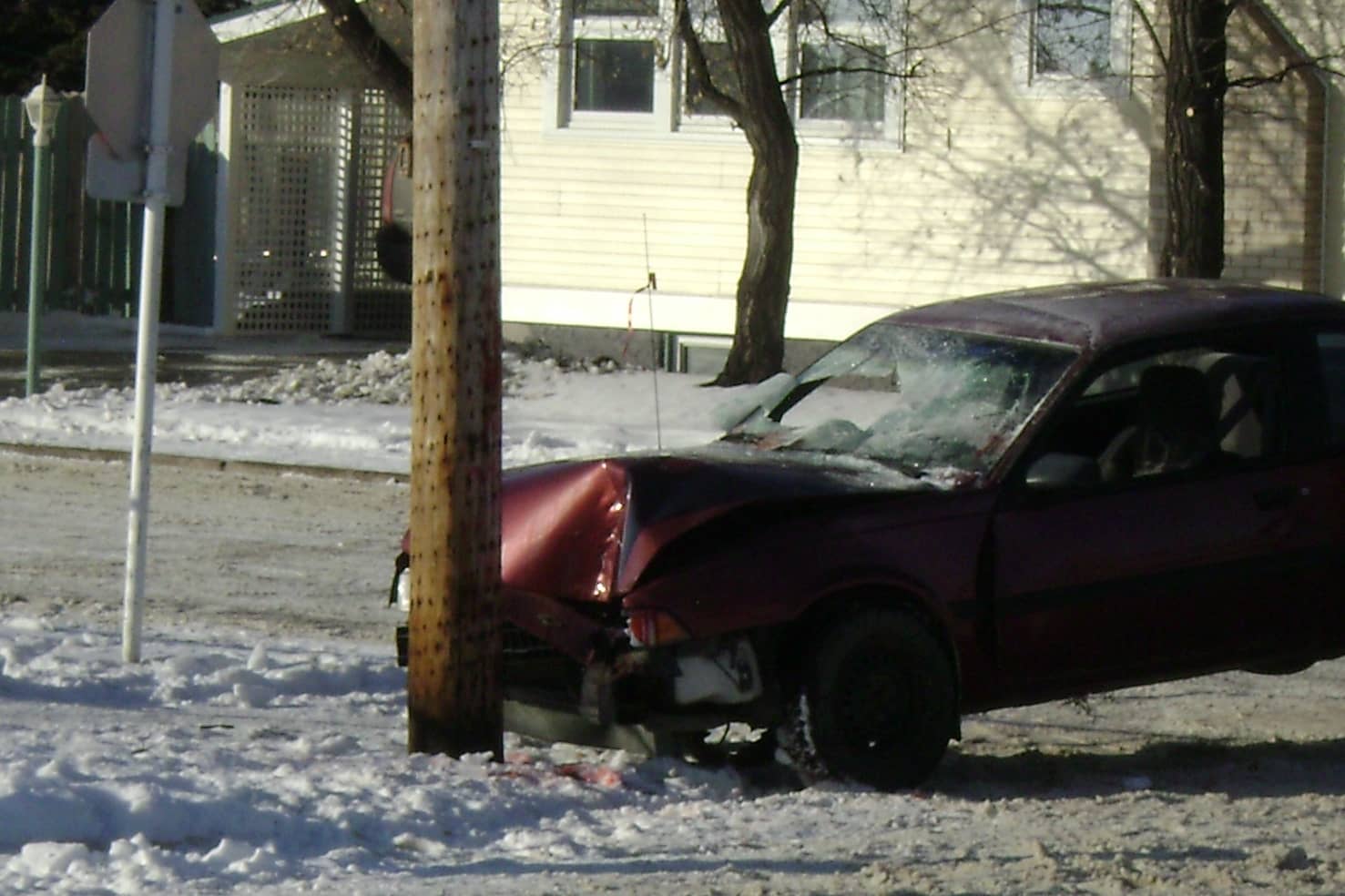}
            \caption{\say{a car is crashed in a snow covered street}}
		\end{subfigure}
	    \begin{subfigure}[b]{0.32\textwidth}
			\includegraphics[scale=.048]{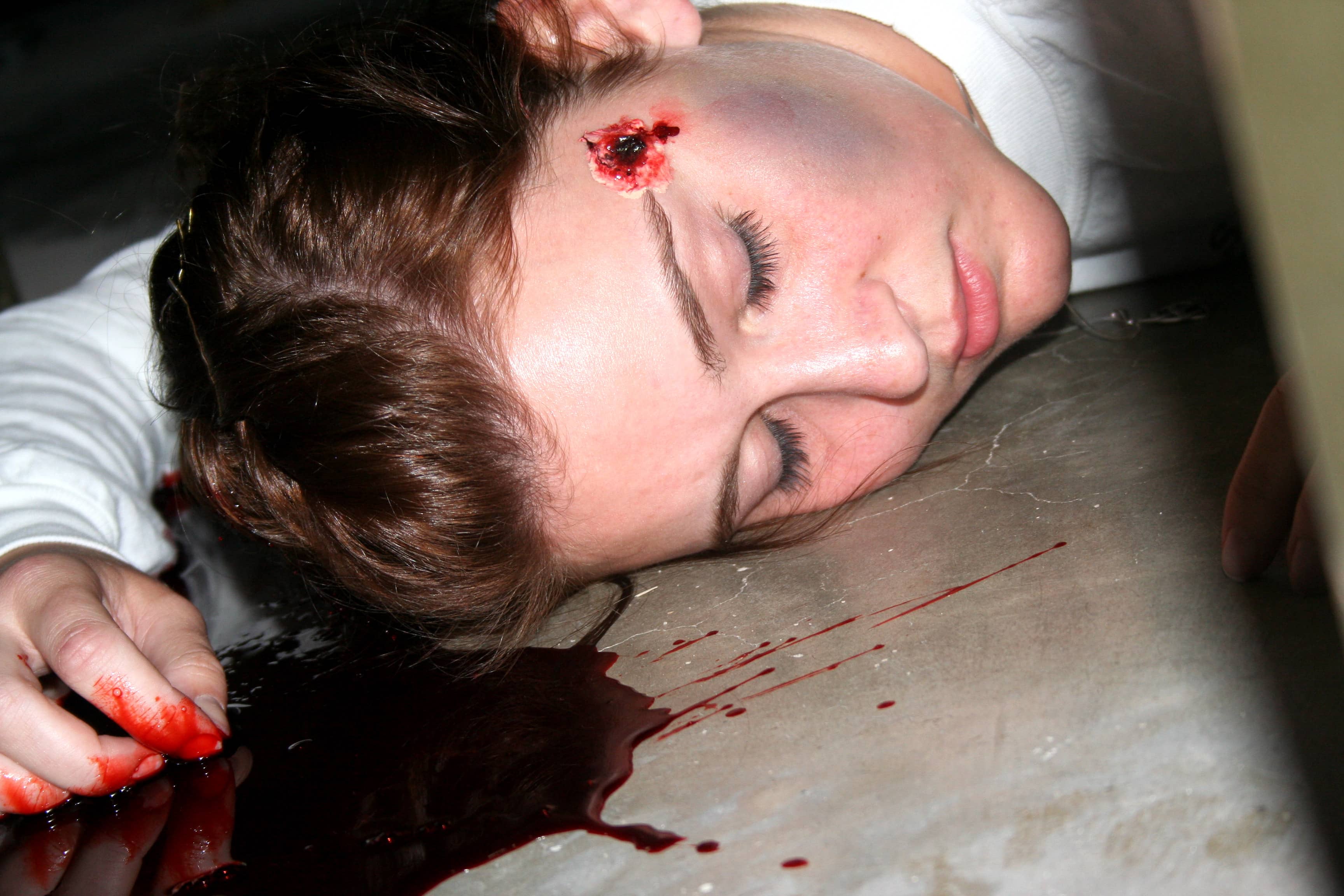}
            \caption{\say{there is a woman with blood on the floor}}
		\end{subfigure}
        \vskip\baselineskip
		~ 
	    \begin{subfigure}[b]{0.32\textwidth}
			\includegraphics[scale=.042]{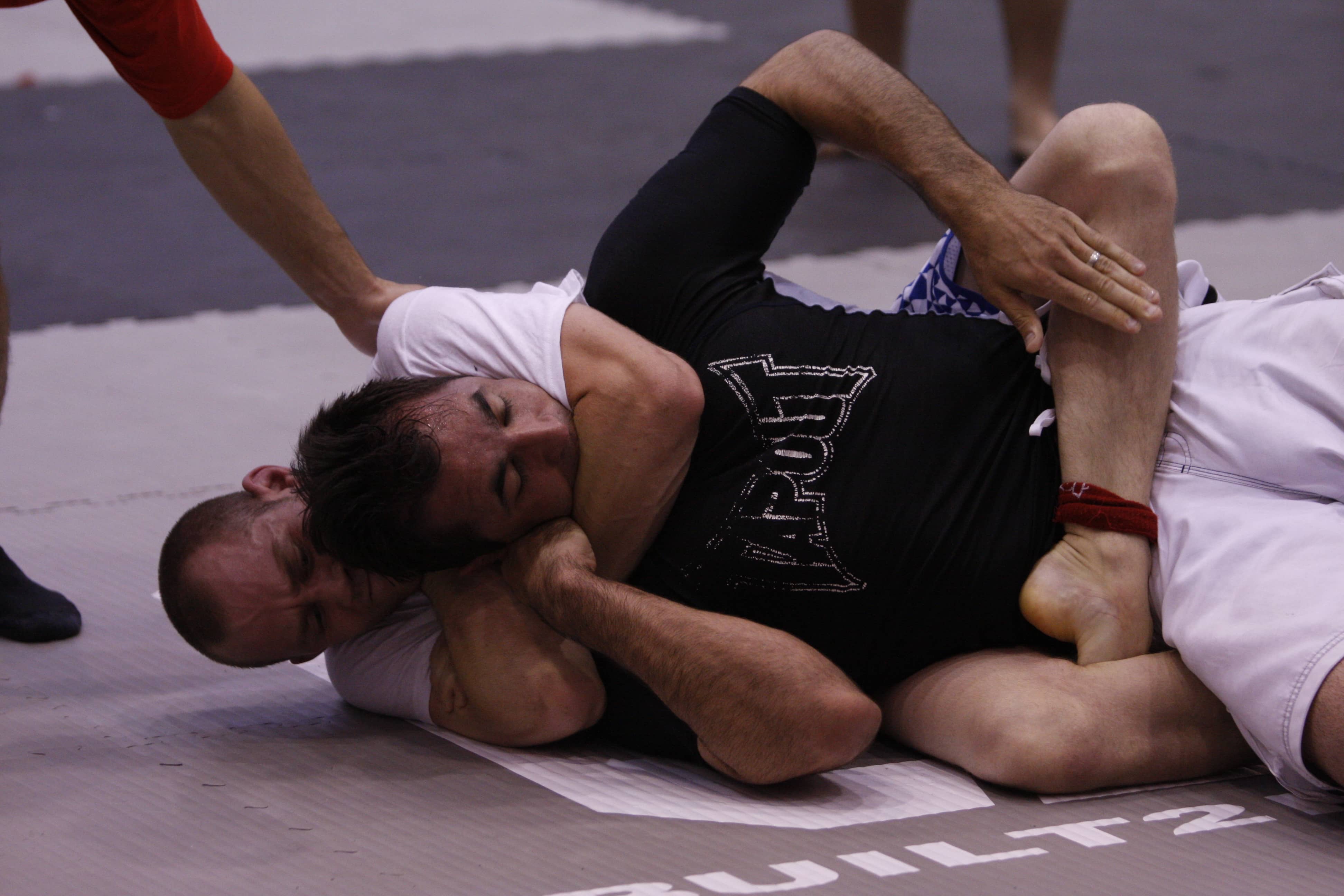}
            \caption{\say{a man is choking another man}}
		\end{subfigure}
		\caption{Correct captions generated by our model when trained on the IAPR-AD dataset.}
        \label{fig:add_results_1}
	\end{figure}

        We obtained a METEOR score of 16.6 in the IAPR and 16.2 in the IAPR-AD test set. This score is better than the reported best for this dataset \cite{Elliott2015} (15.4). However we emphasize that we only used captions that were at most 14 words long. 
        We categorize as positive an image that is not an anomaly, and as negative an image that corresponds to an anomaly.
    Figure \ref{fig:add_results_1} shows correct captions performed by our model when trained on the IAPR and the AD dataset.
    The images here selected, display possible robotic applications which consists of detecting and describing accidents, fires and car crashes.
    In the sub-figure (a) of the same figure, the model correctly describes the anomaly by referring to a \say{gun}.
    Sub-figure (b) correctly displays the situation of a house burning in a concise descriptions consisting of only 4 words.
    Sub-figure (c) made a reference to a car accident \say{a car is crashed} but also to the background \say{snow covered street}.
    Sub-figure (d) also displays in a very concise manner a possible accident by making reference to  \say{woman} \say{blood} \say{floor}.
    Sub-figure (e) shows an example of how our model can be superior to a classification model in regards of anomaly detection.
    Using a simple classification algorithm we could have only detected \say{man}.
    However, our model not only detects two men but also detects the their activity by stating \say{is choking}.
    These two words bring more information about image which could prove beneficial when addressing such situation.
    We report an accuracy of 97 \%.
    The confusion matrix with the number of images is displayed in table 6 and the confusion matrix with ratios is displayed in table 1 and 2. \\

    \begin{center}
        \includegraphics[scale=.4]{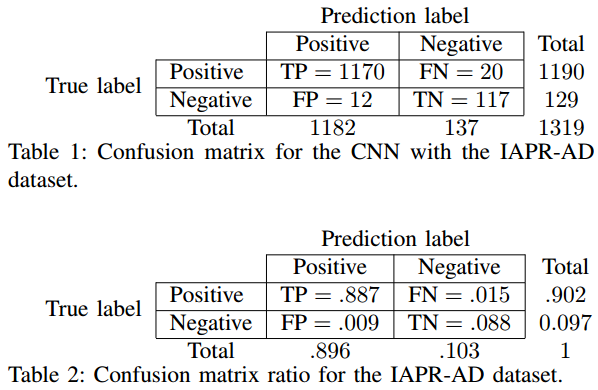}
    \end{center}

    \begin{figure}[H]
    \captionsetup[subfigure]{}
		\centering
		\begin{subfigure}[b]{0.32\textwidth}
			\includegraphics[width=\textwidth]{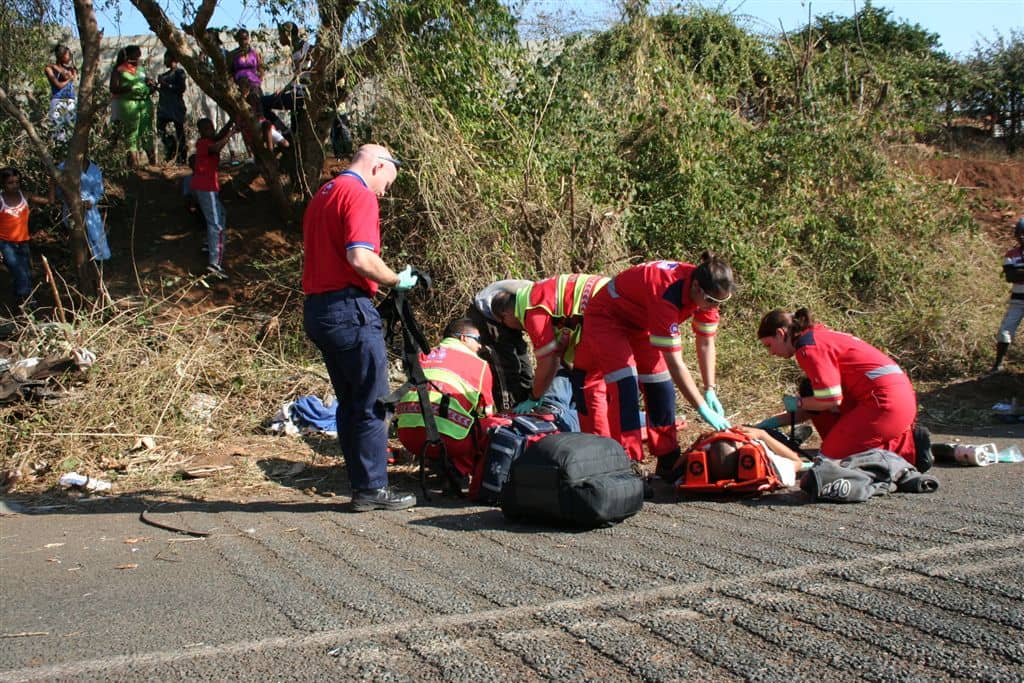}
            \caption{\say{people with red helmets are sitting and cars on a}}
		\end{subfigure}
		\begin{subfigure}[b]{0.32\textwidth}
			\includegraphics[scale=.05]{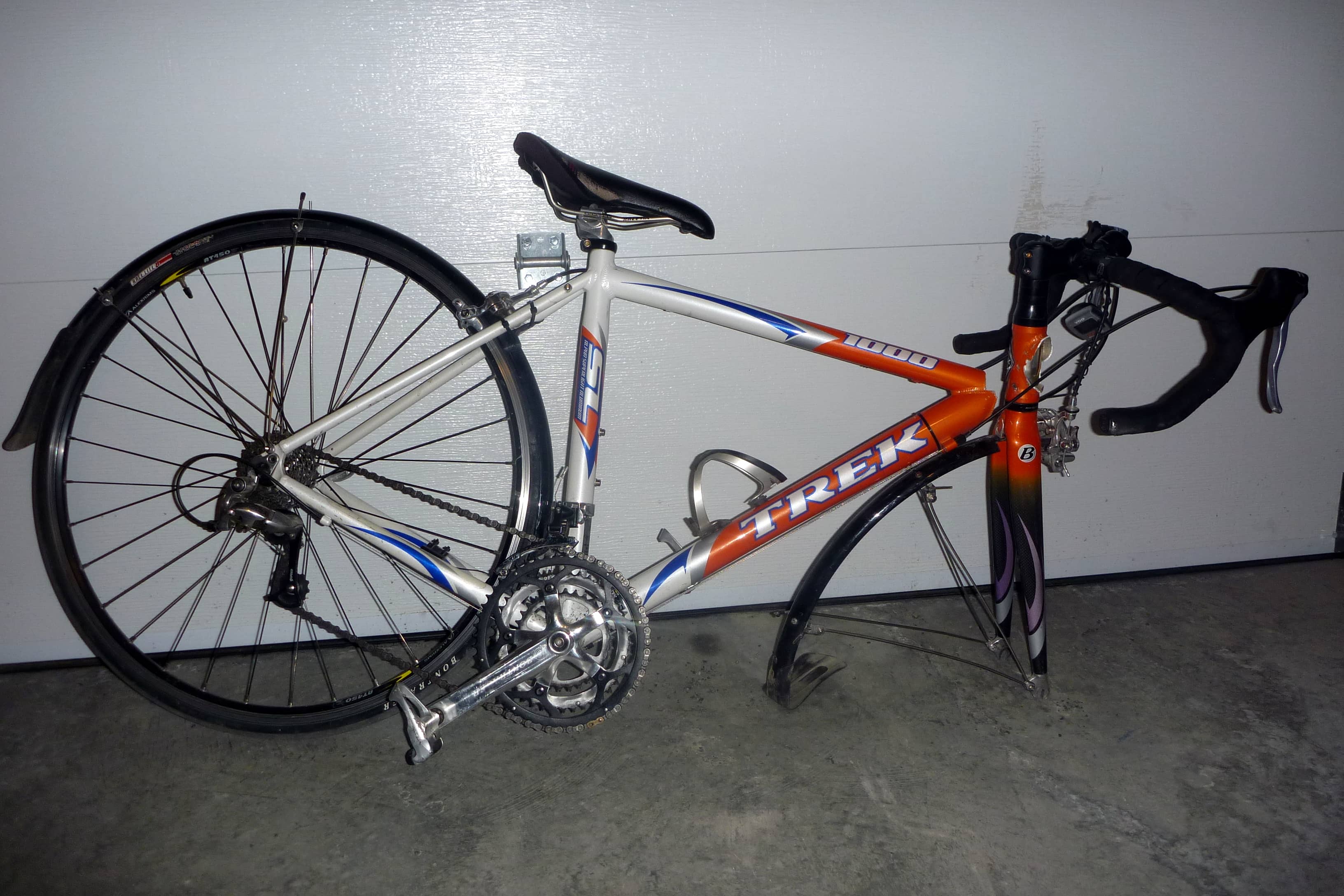}
            \caption{\say{there is a broken window laying on the ground}}
		\end{subfigure}
	    \begin{subfigure}[b]{0.32\textwidth}
			\includegraphics[width=\textwidth]{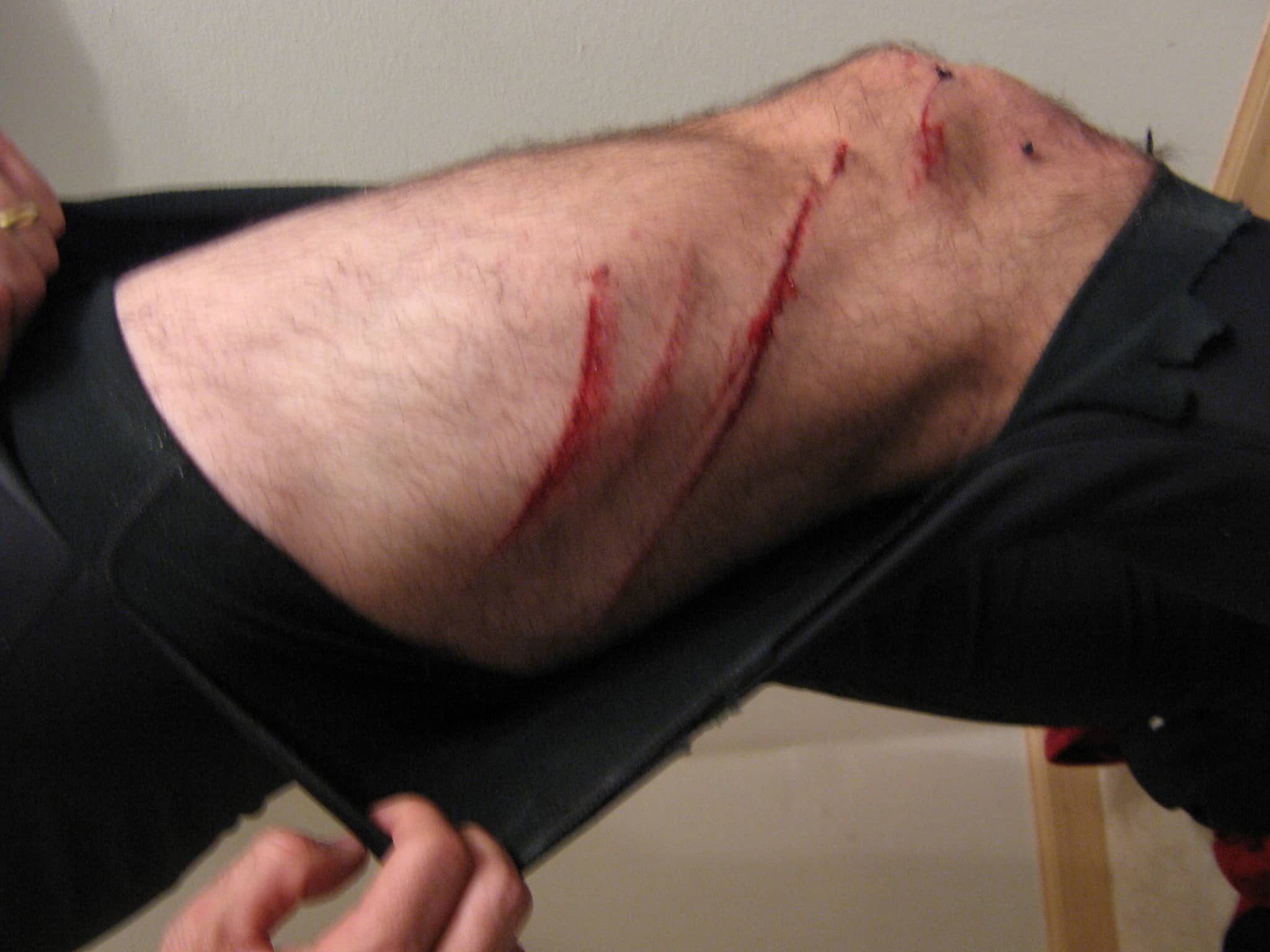}
            \caption{\say{a man showing his right arm in which he has severe injury}}
		\end{subfigure}
        \vskip\baselineskip
		~ 
	    \begin{subfigure}[b]{0.32\textwidth}
			\includegraphics[width=\textwidth]{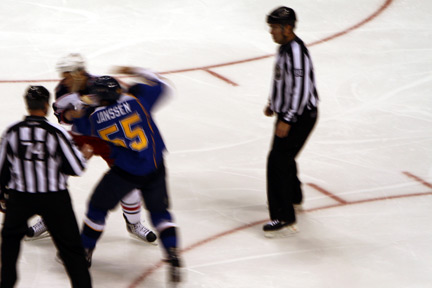}
            \caption{\say{a man is being held by the police}}
		\end{subfigure}
		\caption{Incorrect captions generated by our model when trained on the IAPR-AD dataset.}
        \label{fig:add_results_2}
	\end{figure}

    \newpage
    Figure \ref{fig:add_results_2} displays the incorrect captions performed in the IAPR and AD dataset. 
    Sub-figure (a) from the same image \ref{fig:add_results_2} displays an incorrect classification of \say{helmets} and \say{cars}.
    Sub-figure (b) might refer to the spots found in the ground.
    Sub-figure (d) shows a complicated image in which the leg of man is misclassified as an \say{arm}.
    Sub-figure (e) misclassified the referees by \say{police} officers.
    Caption of non-anomalous images can be seen in Figure \ref{fig:add_results_3} located in the Appendix.

    One important remark that we would like to emphasize; and which has not been thoroughly discussed in any of the image captioning papers, is the correlation between the loss function and METEOR.
    We have found empirically, that in all of our trained models, we obtained better METEOR scores not when the validation loss or accuracy was at its minimum, but several epochs after it.
    We argue that the existence of this discrepancy between the loss function and the METEOR score, is due to the way in which the target variables are represented in all image captioning models.
    The target values are one-hot encodings; consequently, they locate all probability mass in a single word.
    This could also be considered as representing our target values as a hard-evidence.
    However, the METEOR metric uses also a synonym matching module.
    Therefore, there is an underlying symmetry between words, which makes the METEOR metric invariant under specific switching between the one-hot encoding representations.
    This leads us to think that we can relax the hard-evidence constraint and use instead a soft representation of the targets.
    This could be done by distributing uniformly the probability of every single word along its synonyms, or by considering any pre-trained word embedding model and distributing the probability using a Gaussian distribution centered on the original word.
    Testing these hypothesis goes beyond the scope of this research and we leave this for future work.

        \section{Conclusions}
        In response to the current lack of data related to potentially dangerous situations, this paper introduces the anomaly detection (AD) dataset consisting of 1008 captioned images.
        We extended the original NIC model by creating a joint CNN-LSTM + MLP architecture that not only classifies anomalous situations, but also creates a full description in natural language of the anomaly.
        We used our new dataset in combination with the IAPR-2012 dataset in order to train our model and obtain an accuracy of 97 \% and a METEOR score of 16.2.

        Furthermore, we argue that the complete description of the anomaly is fundamental in order to create autonomous systems that are able to attend the dangerous situation appropriately.
        Our final model is able to describe situations involving the following classes: broken windows, injured people, fights, explosions, car accidents, fire accidents, guns and domestic violence.
        \newpage

        \section*{Acknowledgments}
        We gratefully acknowledge the continued support by the b-it Bonn-Aachen International Center for Information Technology and the Hochschule Bonn-Rhein-Sieg.
        The authors would also like to thank Desmond Elliott for his technical input.

        \bibliographystyle{plain}
        \bibliography{anomaly_detection}

        \section{Appendix}
    \begin{figure}[H]
    \captionsetup[subfigure]{}
        \centering
        \begin{subfigure}[b]{0.32\textwidth}
            \includegraphics[width=\textwidth]{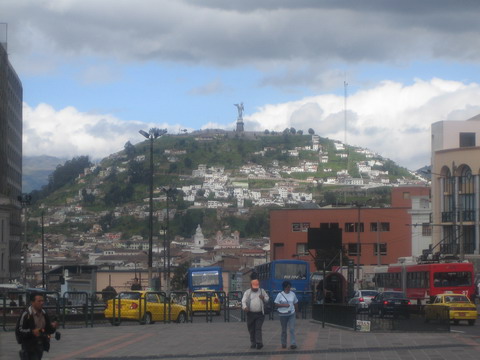}
            \caption{\say{a city with cars and people in the
foreground}}
        \end{subfigure}
        \begin{subfigure}[b]{0.32\textwidth}
            \includegraphics[scale=.32]{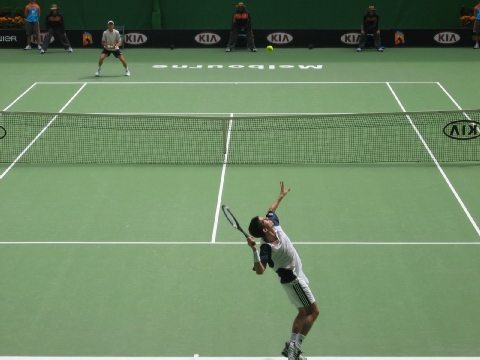}
            \caption{\say{two tennis players on a green hard
court}}
        \end{subfigure}
        \begin{subfigure}[b]{0.32\textwidth}
            \includegraphics[width=\textwidth]{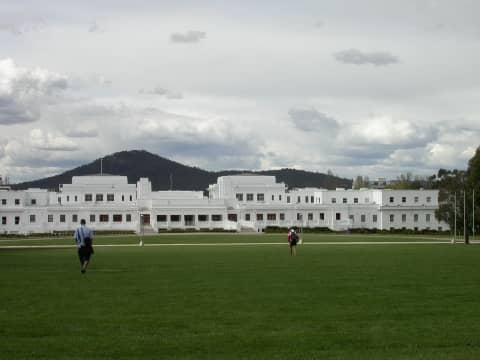}
            \caption{\say{people on a somewhat red square with many trees}}
        \end{subfigure}
        \vskip\baselineskip
        ~ 
        \begin{subfigure}[b]{0.2\textwidth}
            \includegraphics[scale=.28]{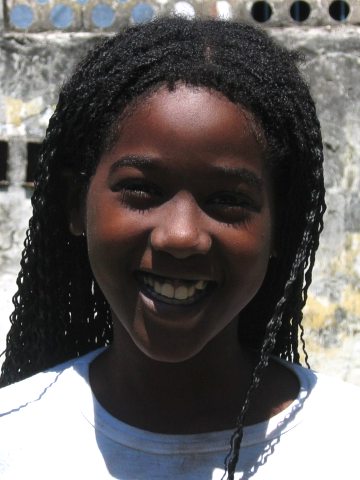}
            \caption{\say{portrait of a black girl with black
hair and a white tee-shirt}}
        \end{subfigure}
        \caption{Generated captions for non-anomalous situations in the IAPR-AD dataset.}
        \label{fig:add_results_3}
    \end{figure}

\end{document}